\documentclass[sigconf]{acmart}
\usepackage{amsmath}

\usepackage{amssymb}
\usepackage{amsthm}
\usepackage{graphicx}
\usepackage{booktabs}
\usepackage{multirow}
\usepackage{url}
\usepackage{balance}

\AtBeginDocument{%
  }

\copyrightyear{2026}
\acmYear{2026}
\setcopyright{cc}
\setcctype{by}
\acmConference[KDD 2026] {Proceedings of the 32nd ACM SIGKDD Conference on Knowledge Discovery and Data Mining V.1}{August 9--13, 2026}{Jeju Island, Republic of Korea.}
\acmBooktitle{Proceedings of the 32nd ACM SIGKDD Conference on Knowledge Discovery and Data Mining V.1 (KDD 2026), August 9--13, 2026, Jeju Island, Republic of Korea}
\acmISBN{979-8-4007-2258-5/2026/08}
\acmDOI{10.1145/3770854.3783927}

\begin{document}

\title{ALF: Advertiser Large Foundation Model for\\Multi-Modal Advertiser Understanding}

\author{Santosh Rajagopalan}
\orcid{0009-0001-7890-7783}
\authornote{Both authors contributed equally to this research.}
\affiliation{%
  \institution{Google}
  \city{Mountain View}
  \state{CA}
  \country{USA}
}
\email{yrj@google.com}

\author{Jonathan Vronsky}\authornotemark[1]
\orcid{0009-0007-1947-8847}
\affiliation{%
  \institution{Google}
  \city{Mountain View}
  \state{CA}
  \country{USA}
}
\email{jvronsky@google.com}

\author{Songbai Yan}
\orcid{0009-0000-0414-7892}
\affiliation{%
  \institution{Google}
  \city{Mountain View}
  \state{CA}
  \country{USA}
}
\email{songbai@google.com}

\author{S. Alireza Golestaneh}
\orcid{0000-0002-3230-2797}
\affiliation{%
  \institution{Google}
  \city{Mountain View}
  \state{CA}
  \country{USA}
}
\email{alirezag@google.com}

\author{Shubhra Chandra}
\orcid{0009-0009-5256-0003}
\affiliation{%
  \institution{Google}
  \city{Mountain View}
  \state{CA}
  \country{USA}
}
\email{shubhrac@google.com}

\author{Min Zhou}
\orcid{0000-0002-3492-600X}
\affiliation{%
  \institution{Google}
  \city{Mountain View}
  \state{CA}
  \country{USA}
}
\email{minzhouis@google.com}

\renewcommand{\shortauthors}{Santosh Rajagopalan et al.}

\begin{abstract}
We present ALF (Advertiser Large Foundation model), a multi-modal transformer architecture for understanding advertiser behavior and intent across text, image, video, and structured data modalities. Through contrastive learning and multi-task optimization, ALF creates unified advertiser representations that capture both content and behavioral patterns. Our model achieves state-of-the-art performance on critical tasks including fraud detection, policy violation identification, and advertiser similarity matching.  In production deployment, ALF demonstrates significant real-world impact by delivering simultaneous gains in both precision and recall, for instance boosting recall by over 40 percentage points on one critical policy and increasing precision to 99.8\% on another. The architecture's effectiveness stems from its novel combination of multi-modal transformations, inter-sample attention mechanism, spectrally normalized projections, and calibrated probabilistic outputs.
\end{abstract}



\maketitle
\section{Introduction}
Online advertising platforms serve as the economic engine of the modern web. A
core challenge in this ecosystem is to accurately and efficiently understand
advertiser intent and behavior. This understanding is critical for several key
applications, including matching users with ads and identifying fraud and
policy violations. Addressing this challenge requires a holistic approach,
processing diverse data types including structured account information (e.g.,
account age, billing details), multi-modal ad creative assets (text, images,
videos), and landing page content. For example, an advertiser might have a
recently created account, have text and image ads for a well-known large
brand, and have had a credit card payment declined once. Although each element could
exist innocently in isolation, the combination strongly suggests a fraudulent
operation.

Developing such a system presents several key challenges:

\begin{itemize}
\item \textbf{Heterogeneous and high-dimensional data:} Advertiser data is
  inherently heterogeneous, encompassing structured (numerical, categorical,
  univalent, and multivalent) and unstructured (text, image, video) data. Each
  data type has different scales, statistical properties, and semantic
  interpretations. Furthermore, the sheer number of features can lead to a
  high-dimensional representation, creating computational and statistical
  challenges. Effectively integrating and modeling these disparate data types
  to create a unified representation, while preserving the informational
  richness of each, requires a sophisticated approach.
\item \textbf{Unbounded sets of creative assets:} Advertisers can have a huge and highly variable number of creative assets. Malicious
  actors may even exploit this by obfuscating fraudulent assets within a large
  volume of innocent ones. The system must be able to process an unbounded and
  potentially adversarial set of assets, identifying malicious content even
  when hidden among numerous benign creatives, without being overwhelmed by the
  volume of data.
\item \textbf{Real-world reliability and trustworthiness:} Predictions made by
  the system directly impact real businesses and users. Therefore, the system
  must be highly reliable and provide well-calibrated confidence scores to
  ensure that the model's outputs are interpretable and actionable. Producing
  calibrated probabilities reduces the overall system-wide operational
  complexity by allowing downstream components to use these probabilities to
  make decisions (for example, should we suspend the account, or ask for identification from this
  advertiser?) without such components needing to be re-tuned after every
  model upgrade.
\end{itemize}


These challenges are not adequately addressed by prior work. Domain-specific solutions, such as those for creative asset classification \cite{hussain2017automatic}, are too narrow and ignore the rich interactions between different data types. On the other hand, recent transformer-based tabular models are not equipped to handle multi-modal inputs, while large multi-modal models are architecturally ill-suited for processing massive, tabular-heavy datasets and are not optimized for the kind of scalable, calibrated predictions required in production environments.

To bridge this critical gap, we present ALF (Advertiser Large Foundation model), a unified transformer architecture for real-world advertiser understanding. ALF creates a holistic understanding of advertisers by jointly modeling their structured data and multi-modal assets. Our primary contributions, which directly address the challenges of heterogeneity, scale, and reliability, are as follows:

\begin{itemize}
\item \textbf{An Efficient Holistic Architecture for Heterogeneous Data:} We propose an efficient, holistic architecture featuring a specialized transformer. Its unified encoding strategy performs an early fusion where structured data and multi-modal embeddings are mapped into a shared space, enabling a dual-attention mechanism to learn deep cross-modal interactions. This mechanism includes a scalable, projection-based inter-sample attention that overcomes the hidden dimension and input length scaling limitations of prior work \cite{somepalli2021saint}, allowing the model to effectively learn from interactions across large advertiser batches.

\item \textbf{Scalable Handling of Unbounded Creative Assets:} To efficiently manage a large and variable number of creatives, we propose to process pre-computed asset embeddings instead of raw bytes and uses a top-k selection mechanism. This ensures robust and scalable performance, particularly in adversarial scenarios.

\item \textbf{Trustworthy Predictions for Real-World Deployment:} We integrate Spectrally-Normalized Neural Gaussian Process (SNGP) heads during fine-tuning. This produces well-calibrated probabilistic outputs, which are crucial for high-stakes decision-making and simplify integration with downstream production systems.
\end{itemize}

Our experiments show ALF significantly outperforms a heavily-tuned production baseline while also performing strongly on public benchmarks. In production, ALF delivers substantial and simultaneous gains in precision and recall, boosting recall by over 40 percentage points on one critical policy while increasing precision to 99.8\% on another. This performance lift is driven by ALF's unique ability to integrate multi-modal creative content with tabular features, a task where traditional architectures falter.

The remainder of this paper is organized as follows. Section \ref{sec:related}
discusses related work in multi-modal learning and advertiser understanding. Section \ref{sec:problem} formulates the problem space.
Section \ref{sec:model} describes our model architecture in detail. Section
\ref{sec:training} presents our training methodology, while Section
\ref{sec:experiments} provides comprehensive experimental results and
analysis. Finally, Section \ref{sec:conclusion} concludes with a discussion of future work.

\section{Related Work}
\label{sec:related}

\subsection{Multi-modal Learning}
Recent work in multi-modal learning has demonstrated the effectiveness of transformer architectures in combining different data types. Models like CLIP \cite{radford2021learning} and DALL-E \cite{ramesh2021zero} have shown strong performance on joint image-text tasks through contrastive learning approaches. More recently, large generative multi-modal models, such as Flamingo 
\cite{alayrac2022flamingovisuallanguagemodel}, FLAVA \cite{singh2022flava}, VLMo \cite{bao2022vlmounifiedvisionlanguagepretraining},  OmniVL \cite{wang2022omnivl}  introduced a unified foundation model for visual and language tasks, showing effective zero-shot transfer across multiple tasks. However, such large models are computationally expensive at our scale because they operate on limited number of raw multi-modal data (e.g., image pixels) rather than dense embeddings. Our approach, in contrast, leverages efficient, pre-computed embeddings as inputs to a specialized predictive architecture.

In the advertising domain, \citet{hussain2017automatic} developed techniques for creative asset classification, while \citet{rayavarapu2022multimodal} investigated transformers for detecting bad quality ads. However, these efforts primarily focused on isolated modalities or specific tasks, unlike our holistic approach that integrates all available advertiser signals.

\subsection{Transformers for Structured Data}
Traditional approaches to structured data typically rely on gradient-boosted decision trees (GBDT) \cite{ke2017lightgbm, chen2016xgboost}. However, recent work has shown promising results using transformers for tabular data. TabTransformer \cite{huang2020tabtransformer} demonstrated strong performance through categorical embedding learning, while SAINT \cite{somepalli2021saint} showed particular effectiveness through intersample attention mechanisms.

FT-Transformer \cite{gorishniy2021revisiting} provided a comprehensive analysis of transformer architectures for tabular data, showing competitive performance against traditional approaches. Recent work by \citet{yoon2020vime} on self-supervised learning for tabular data has shown the importance of handling missing values and heterogeneous features. Work on Non-Parametric Transformers (NPT) \cite{kossen2022selfattentiondatapointsgoingindividual}, in addition to SAINT \cite{somepalli2021saint}, shows the importance of cross-attention across data points in transformers for tabular data.

Our work builds on this line of research by adapting the transformer architecture with cross-attention to the specific challenges of advertiser modeling. Specifically, we extend the model's ability to handle both structured and unstructured multi-modal data of large volume and dimensionality, and introduce techniques for generating calibrated predictions, critical for real-world deployment.

A recent line of work has explored the application of large language models (LLMs) to numerical data using custom tokenization schemes (e.g., P10) \cite{song2025omnipredlanguagemodelsuniversal}. Similar to the large multi-modal models mentioned previously, these methods suffer from significant cost and scaling issues.

\subsection{Advertiser Understanding}
Prior work on advertiser understanding has focused on specific aspects of the advertising ecosystem. \citet{Dave2012ClickSpam} developed methods for click fraud detection using temporal patterns and behavioral analysis. Landing page and image categorization for display ads has been explored by \citet{kae2011categorization}.

Deep learning approaches to ad fraud detection have been investigated by \citet{gopali2024performancesequentialdeeplearning} using sequential models and \citet{xu2024revisitinggraphbasedfrauddetection} through graph-based methods. Policy violation detection in online advertising has been studied by \citet{mittal2023usingfoundationmodelsdetect} using Foundation models and \citet{qu2024multitaskcnnbehavioralembedding} through multi-task learning approaches.

Our work differs from these approaches by providing a unified model that jointly processes all advertiser-related signals, including structured data, creative assets, and behavioral patterns. This unified approach, purpose-built for Advertiser understanding, allows for better feature interaction and knowledge transfer across tasks.
\section{Problem Formulation}
\label{sec:problem}

In this paper, we consider an advertiser understanding task, where each example
is a snapshot of an advertiser $x$ drawn from a distribution $D$, consisting
of:
\begin{itemize}
    \item Structured features $s_x \in \mathbb{R}^d$ including categorical and
      numerical, and possibly multivalent, account attribute and historical
      performance metrics;
    \item Text content $T_x = \{t_1, ..., t_n\}$ including ad text, keywords,
      and landing page text;
    \item Image content $I_x = \{i_1, ..., i_m\}$ from ad creatives, landing
      pages (and landing page screenshots);
    \item Video content $V_x = \{v_1, ..., v_k\}$ from video ads.
\end{itemize}

We are given $m$ tasks: for the $t$-th task, we assume that there is a ground
truth label $y^t$ drawn from an unknown distribution $D_t(Y\mid X=x)$. We are
given a loss function $l_t$ and we would like to find a model $m$ that minimizes
$\mathbb{E}_{D_t}[l_t(m(x), y^t)]$. In this paper, we focus on multiclass
classification tasks, but our method can be easily generalized to broader tasks.

\section{Model Architecture}
\label{sec:model}

\begin{figure*}[t]
\centering
    \includegraphics[width=0.6\textwidth]{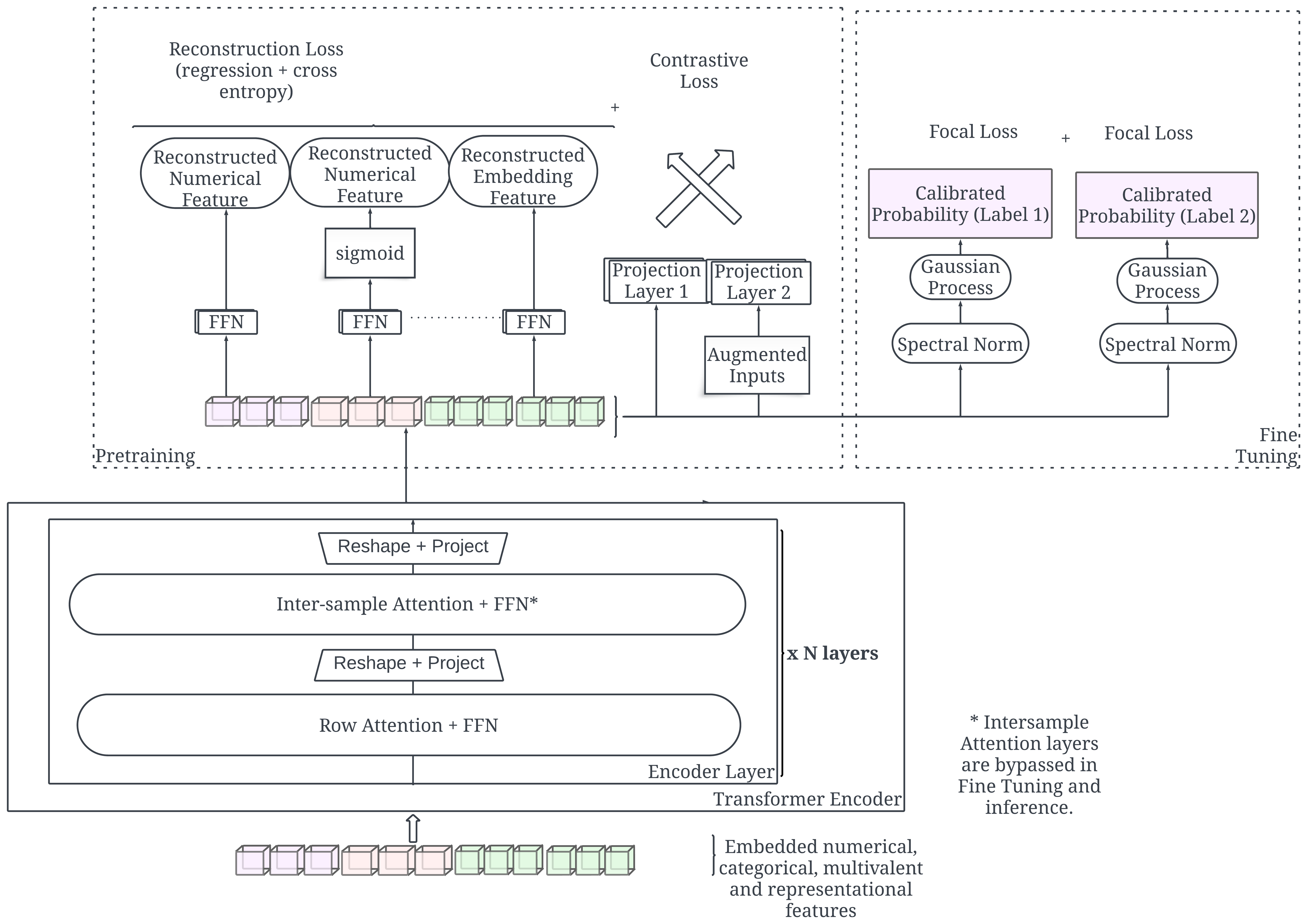}
    \caption{ALF model architecture showing the multi-modal encoders, dual
    attention mechanisms, and output heads.}
    \Description{ALF model consists of a multi-modal encoder to map input
    numerical, multivalent, and representational features into the same
    embedding space, a transformer encoder with dual attention mechanisms, and
    a set of output heads for pretraining and finetuning tasks.}
    \label{fig:architecture}
\end{figure*}

The ALF architecture, shown in Figure~\ref{fig:architecture}, constructs a
unified advertiser representation from heterogeneous, multi-modal data
sources. It begins by projecting heterogeneous raw features -- including
structured numerical/categorical data (both univalent and multivalent), and
unstructured text, image, and video—into a common embedding space using a
novel encoding approach. This enables ALF to handle a variety of advertiser
data in a scalable manner. A novel dual-attention transformer encoder,
combining self-attention with scalable inter-sample attention, then processes
these embeddings to produce a robust advertiser embedding. During
pre-training, we apply FFN and projection layers over the advertiser embedding
to train ALF via self-supervised reconstruction and contrastive learning
tasks. For downstream prediction tasks in fine-tuning, ALF employs
task-specific, Spectrally-Normalized Neural Gaussian Process (SNGP) heads.
These SNGP heads provide calibrated probabilistic outputs, crucial for
reliable, risk-aware decision-making in advertising. We explain each component
in the following two sections.

\subsection{Input Processing}\label{subsec:input}
\begin{figure}[t]
\centering
    \includegraphics[width=0.4\textwidth]{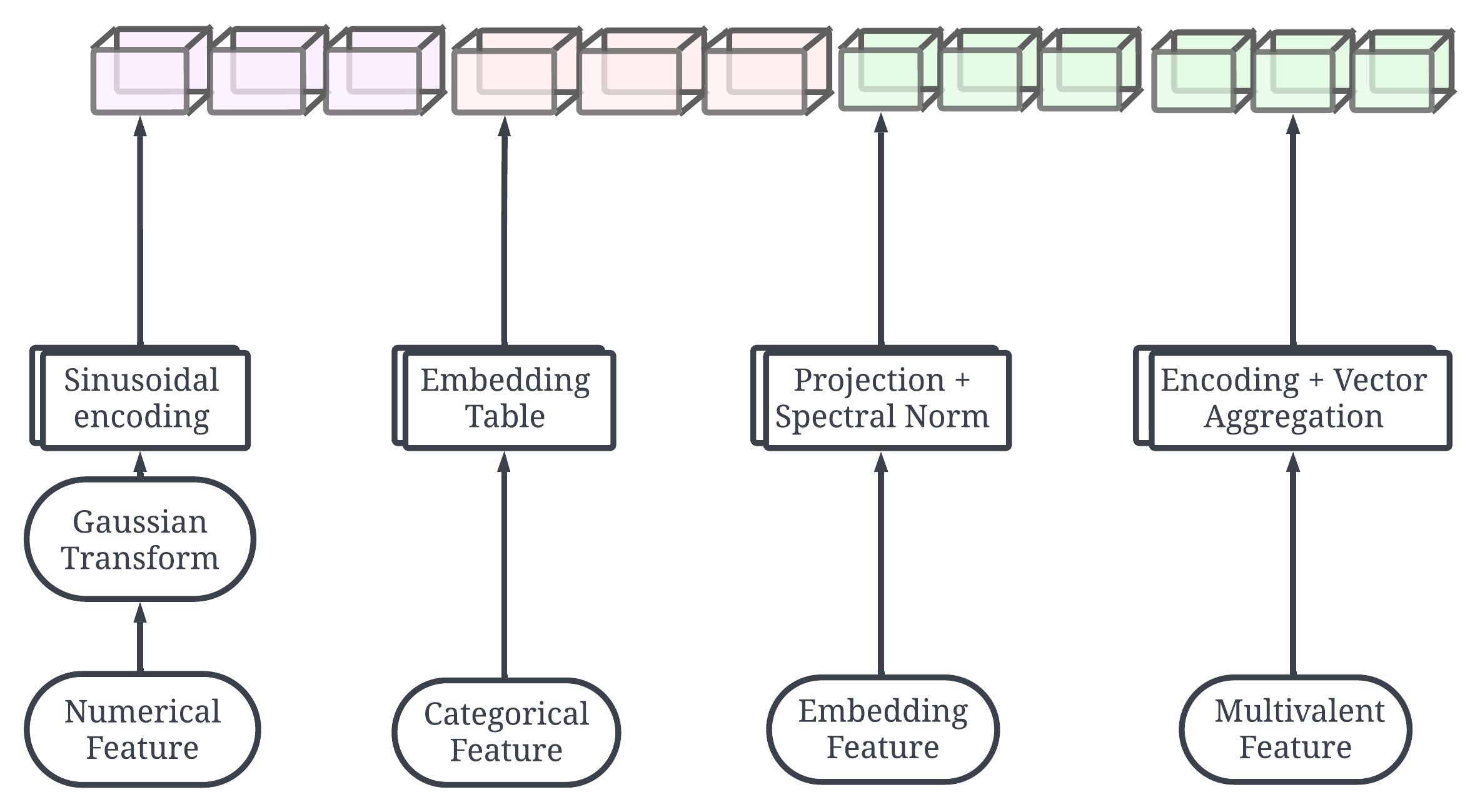}
    \caption{ALF input processing for each feature type.}
    \Description{In ALF, we map heterogeneous numerical, categorical,
    embedding, and multivalent embedding features into a shared embedding
    space, which are then fed into the Transformer trunk.}
    \label{fig:input_processing}
\end{figure}

We have a heterogeneous feature space with structured categorical and
numerical features, and possibly unbounded number of advertiser ad texts,
images, and videos. Common approaches to model such heterogeneous features and
the interactions among them are either through gradient boosted decision trees
\cite{ke2017lightgbm, chen2016xgboost} or variants of factorized machines
\cite{wang2021dcn}, but they often struggle with the scale and complexity of
modern advertising data, or limit cross-modal interactions. In contrast, ALF
leverages a Transformer architecture backbone. To harness this power, we
introduce a novel encoding approach (depicted in
Figure~\ref{fig:input_processing}) that transforms all input modalities into a
unified, shared embedding space. This scalable encoding is essential for the
downstream dual-attention mechanism and forms the foundation of our coherent
customer representation.

It is important to note that positional encoding is not used at this stage.
The input to the transformer is treated as an unordered set of features, as the
structured data and asset embeddings do not have an inherent sequential order.
Any positional information relevant to sequence-based modalities like text or
video is handled by the upstream models that generate the asset embeddings.

\vspace{2em}

\subsubsection{Structured Data Encoding}
Following \citet{somepalli2021saint}, we map each individual structured feature
into $d$ dimensional space. For numeric features, unlike
\cite{somepalli2021saint} which uses multilayer perceptrons to map
1-dimensional scalars into $d$ dimensional space, here we directly encode them
using sinusoidal positional encoding. This choice offers a good balance of non-linear transformation capability and
parameter efficiency, which is crucial for handling a large number of
numerical features:
\begin{align}
    \text{SE}(x, 2i) & = \sin(x \cdot f_i \cdot \pi) \\
    \text{SE}(x, 2i + 1) & = \cos(x \cdot f_i \cdot \pi)
\end{align}
where $f_i$ ($i=1,\dots, d$) are learned frequency parameters. This reduces the
number of parameters for encoding and helps scale up the number of features we
could support, while our empirical evaluation shows comparable performance
to MLPs.

For categorical features, we use learnable embeddings
$E_c \in \mathbb{R}^{|V| \times d}$ where $|V|$ is the vocabulary size.
Embeddings for multi-label features are combined through summation:
\begin{equation}
    e_{\text{cat}} = \sum_{i \in C} E_c[i]
\end{equation}

\subsubsection{Multi-modal Encoders}
Text, image and video contents are converted to embeddings using
transformer-based encoders. Here we adopt multi-lingual LaBSE
\cite{feng2020labse} for texts and GRAPH-RISE \cite{Juan2018GraphRISE} for
images and videos, but we expect any encoder trained to maximize the entropy
of the generated embeddings \cite{chakraborty2024improving, liang2021learning}
will work equivalently.
\begin{align}
    e_{\text{text}} &= \text{LaBSE}(t) \in \mathbb{R}^{d_t} \\
    e_{\text{image}} & = \text{ImageEncoder}(i) \in \mathbb{R}^{d_i} \\
    e_{\text{video}} &= \text{VideoEncoder}(v) \in \mathbb{R}^{d_v}
\end{align}

These embeddings are then mapped to $d$-dimensional space (same as the
encoding for structured data) via multilayer perceptrons.

\subsubsection{Selecting Advertiser Assets and Landing Pages}
An advertiser can have a potentially unbounded number of assets, such as
images, videos, and landing pages. Including all of them would increase the
cost of both training and inference. Instead, we select the top-k most
representative assets of an advertiser. The top-k selection method employed
depends on the specific application and could include selecting cluster centroids and
outliers, or selecting based on recency, user engagement, and content type. These could be supplemented
with a random selection of assets as well to give a comprehensive understanding
of the advertiser and their intent.

\subsection{Feature Selection}
The number of structured features for an advertiser can be very high,
encompassing a wide range of metadata, attributes, and derived metrics. To
ensure that our model focuses on the most relevant signals and to maintain
computational efficiency, we employ a preliminary feature selection step before
the main training pipeline. This step uses a backward elimination approach,
where a Random Forest model trained with the Yggdrasil Decision Forests (YDF)
library
\cite{GuillameBert2022YggdrasilDF}\footnote{\url{https://github.com/google/yggdrasil-decision-forests}}
iteratively prunes the least important features based on their importance
scores.

This feature selection process is not part of the core ALF training pipeline
but is a crucial pre-processing step that allows us to work with a more
manageable and informative set of structured features. The final selected
features are then used as input to the main ALF model, where they are combined
with the multi-modal embeddings for the dual-attention transformer.

\subsection{Scalable Dual Attention Transformer Architecture}
\label{subsec:transformer}

ALF uses a modified transformer architecture with dual attention mechanisms.
Define the attention function:

\begin{equation}
\text{Attn}(X; W_Q, W_K, W_V) = \text{softmax}\left(\frac{(XW_Q)(XW_K)^T}{\sqrt{d}}\right)(XW_V)
\end{equation}

Recall that in standard row attention, for a batch of examples
$X \in \mathbb{R}^{B\times N\times d}$ of $B$ samples, we directly apply the
attention function on $X$ with three trainable matrices
$W_Q, W_K, W_V \in \mathbb{R}^{d\times d}$. As a result,
$Q, K, V = \text{Attn}(Q, K, V)$ are all tensors in
$X \in \mathbb{R}^{B\times N\times d}$.

Inspired by \cite{somepalli2021saint}, we apply inter-sample attention within
a batch to incorporate information about the distribution for each individual
feature, which would be helpful if a feature is missing or noisy for one
sample. We propose here a more scalable version of inter sample attention
(ISA) to address issues we observed going to higher dimensions, features and
batch sizes.

After the initial attention and FFN layers, we reshape and project the
$Nd$-dimensional embeddings to a lower dimension $d'$ using a learnable
projection matrix $P \in \mathbb{R}^{Nd \times d'}$ to improve computational
efficiency:

\begin{align}
X_{\text{reshaped}} &= \text{reshape}(X, (1, B, Nd)) \\
X_{\text{proj}} &= X_{\text{reshaped}}P
\end{align}

where $X_{\text{proj}} \in \mathbb{R}^{1 \times B \times d'}$. This projection
reduces the computational complexity of both the subsequent attention
operation from $O(B^2Nd)$ to $O(B^2d')$ and the FFN layer from $O(BNd^2)$ to
$O(Bd'^2)$, making it more scalable for larger batch sizes and feature
dimensions.

Next, we apply the attention mechanism on the projected embedding
$X_{\text{proj}}$. Note that here the second dimension of $X_{\text{proj}}$ is
of size $B$, so the attention here is inter-sample within the training batch
to supplement the embedding vector with information from other examples in the
batch for missing or noisy features.

\begin{align}
A_{\text{IS}} &= \text{Attn}(X_{\text{proj}}, W'_Q, W'_K, W'_V)
\end{align}

The projected representations then pass the FFN layer while maintaining the
reduced dimensionality $d'$. After the FFN, we project back to the original
dimension using a learnable matrix
$P_{\text{restore}} \in \mathbb{R}^{d' \times Nd}$:

\begin{align}
X_{\text{restored}} &= \text{FFN}(A_{\text{IS}})P_{\text{restore}} \\
X_{\text{final}} &= \text{reshape}(X_{\text{restored}}, (B, N, d))
\end{align}

This projection-based architecture substantially improves computational
efficiency while preserving the model's ability to capture inter-sample
relationships.

As shown in Figure \ref{fig:architecture}, we stack $N$ such layers as our
Dual Attention Transformer Encoder.

Note that inter-sample attention layers are bypassed in the fine tuning and
inference stage to avoid cross-advertiser information leakage.

\subsection{Spectral Normalization}
All projection layers use spectral normalization to keep Lipschitz constants
bounded:

\begin{equation}
W_{\text{SN}} = \frac{W}{\sigma(W)}
\end{equation}

where $\sigma(W)$ is the spectral radius (largest singular value) of matrix
$W$, computed efficiently using power iteration \cite{miyato2018spectral}.
This normalization ensures that the Lipschitz constant of each layer remains
bounded, which:
\begin{itemize}
\item Stabilizes training by preventing extreme variations in gradients
\item Makes the embedding space more isotropic by avoiding dimension collapse
\item Improves robustness of the learned representations against input
perturbations
\end{itemize}

The power iteration method approximates $\sigma(W)$ using:
\begin{align}
u_{t+1} &= \frac{W^\top u_t}{\|W^\top u_t\|_2} \\
v_{t+1} &= \frac{W u_{t+1}}{\|W u_{t+1}\|_2} \\
\sigma(W) &\approx u_{t+1}^\top W v_{t+1}
\end{align}
where $u_t, v_t$ are the left and right singular vectors corresponding to the
largest singular value.

\section{Training}
\label{sec:training}
The model is trained in two stages: a pre-training stage with self-supervised representation learning and a fine-tuning stage for supervised applications.

\subsection{Pre-training}
The model is pre-trained on over 100 million advertiser snapshots to learn a
robust representation via a contrastive learning task and a reconstruction
task after data augmentation.

Specifically, we employ two augmentation strategies:
\begin{itemize}
    \item CutMix\cite{cutmix2019}: Combines different samples in input space
    \begin{equation*}
        x' = M \odot x_i + (1-M) \odot x_j
    \end{equation*}
    where $x_i$ and $x_j$ are two samples in the input space, and $M$ is a
    binary mask drawn from a Bernoulli distribution $B(0.2)$.

    \item MixUp\cite{DBLP:journals/corr/abs-1710-09412}: Interpolates samples
      in latent space
    \begin{equation}
        h' = \alpha h_i + (1-\alpha)h_j
    \end{equation}
    where $h_i$ and $h_j$ are two embedding vectors for two samples after the
    feature encoder but before the transformer encoder, and $\alpha$ is a
    weight parameter that can be tuned.
\end{itemize}

After data augmentation, we train the model to minimize a linear combination
of two losses:
\begin{itemize}
\item Reconstruction losses: For augmented input $\tilde{x}$, we use several
  reconstruction objectives for each of the feature type:
\begin{align}
\mathcal{L}_{\text{num}} &= \|f_{\text{num}}(g(\tilde{x})) - x_{\text{num}}\|^2 \\
\mathcal{L}_{\text{ce}} &= -\sum_i x_i \log(f_{\text{ce}}(g(\tilde{x}))_i) \\
\mathcal{L}_{\text{mcat}} &= -\sum_i \sum_j x_{ij} \log(f_{\text{mcat}}(g(\tilde{x}))_{ij}) \\
\mathcal{L}_{\text{emb}} &= \|f_{\text{emb}}(g(\tilde{x})) - x_{\text{emb}}\|^2 \\
\mathcal{L}_{\text{memb}} &= \sum_i \|f_{\text{memb}}(g(\tilde{x}))_i - x_{\text{memb},i}\|^2
\end{align}
where $f_{\text{num}}$, $f_{\text{ce}}$, $f_{\text{mcat}}$, $f_{\text{emb}}$,
$f_{\text{memb}}$ and $g$ are the corresponding decoder and encoder functions,
$\mathcal{L}_{\text{num}}$ is the MSE Loss on the reconstruction of numerical
features, $\mathcal{L}_{\text{ce}}$ is the Cross-Entropy Loss for categorical
feature reconstruction, $\mathcal{L}_{\text{mcat}}$ is the multivalent
Categorical Loss for multi-label scenarios, $\mathcal{L}_{\text{emb}}$ is the
MSE loss for pre-trained embeddings, and $\mathcal{L}_{\text{memb}}$ is the MSE
loss for multivalent embeddings where each feature can have multiple embedding
vectors. These reconstruction losses, in conjunction with the encoder $g$ and their respective 
decoders ($f_{\text{num}}$, $f_{\text{ce}}$, etc.), ensure that the encoder is information-preserving
by accurately reconstructing the various input feature types.

\item Contrastive loss: We use InfoNCE loss between original and augmented
  samples:
\begin{equation}
\mathcal{L}_{\text{con}} = -\log \frac{\exp(s(x,x^+)/\tau)}{\sum_{x^-} \exp(s(x,x^-)/\tau)}
\end{equation}
where $x$ is the original input, $x^+$ is the augmented input of $x$, $x^-$ is
the augmentation of another input drawn within the batch, $s(\cdot,\cdot)$ is
cosine similarity and $\tau$ is a temperature parameter.
\end{itemize}
The total loss is then computed as:
\begin{equation}
\begin{split}
\mathcal{L}_{\text{total}} = \alpha_{\text{num}} \mathcal{L}_{\text{num}} +
\alpha_{\text{ce}} \mathcal{L}_{\text{ce}} + \alpha_{\text{mcat}} \mathcal{L}_{\text{mcat}}  +  \\
\alpha_{\text{con}} \mathcal{L}_{\text{con}} + \alpha_{\text{emb}} \mathcal{L}_{\text{emb}} + \alpha_{\text{memb}} \mathcal{L}_{\text{memb}}
\end{split}
\end{equation}
where $\alpha_i$ are the corresponding scaling factors for each loss term.

\subsection{Fine-tuning}
After pre-training, we fine-tune the model for specific downstream tasks using
a Spectral-normalized Neural Gaussian Process (SNGP) layer
\cite{liu2020simple}\footnote{
\url{https://keras.io/examples/keras_recipes/uncertainty_modeling_with_sngp/}}.
SNGP combines spectral normalization with a random feature approximation of a
Gaussian process, enabling both calibrated uncertainty estimation and robust
predictions far from the training distribution.

For each task $t$, we use focal loss \cite{lin2017focal} to handle class imbalance:
\begin{equation}
\mathcal{L}_{\text{focal},t} = -\sum_{i=1}^N (1-p_i)^\gamma y_i \log(p_i)
\end{equation}
where $p_i$ is the model's predicted probability for the correct class, $y_i$ is
the ground truth label, and $\gamma$ is the focusing parameter that reduces the
relative loss for well-classified examples.

A potential concern with focal loss is its tendency to degrade model
calibration. In our framework, this is mitigated by the SNGP head, which is the
primary component responsible for calibration. SNGP yields reliable uncertainty
estimates, making it particularly effective for out-of-distribution detection
\cite{liu2020simple}. We find the two methods to be complementary: SNGP ensures
that model outputs are well-calibrated, while focal loss improves accuracy on
in-distribution data by addressing severe class imbalance. This combination
allows us to achieve both high accuracy on our imbalanced dataset and reliable
calibration\cite{ye2023efficientuncertaintyestimationgaussian}.

The total fine-tuning loss is then computed as sum across all tasks:
\begin{equation}
\mathcal{L}_{\text{finetune}} = \sum_{t} \mathcal{L}_{\text{focal},t}
\end{equation}

\subsection{Inference}
During inference, the inter-sample attention layers are bypassed.
This ensures that predictions for a single advertiser are self-contained and
can be made without requiring a batch of other advertisers.
This architectural choice allows for efficient,
on-demand evaluation of individual advertisers in a production environment.
The additional runtime overhead of ALF compared to simpler models is primarily
due to the transformer backbone itself, as the conversion of creative assets
into embeddings occurs asynchronously and does not impact online inference
latency.

\section{Experiments} \label{sec:experiments}
\subsection{Evaluation with Public Datasets}
We validate ALF against state-of-the-art competitors using a suite of public benchmark datasets. However, since public data combining structured and multi-modal inputs is not readily available, these benchmarks primarily serve to demonstrate baseline competitiveness. ALF’s core strengths—specifically its ability to integrate multi-modal signals at scale—are comprehensively evaluated on proprietary production data in the next subsection.

\paragraph{Datasets} 
We selected a set of public datasets that are standard benchmarks in the field of tabular deep learning. We considered only datasets with large number of examples and/or features. The details of these datasets are summarized in Table 1, with their public source links provided in Appendix Table 2.


\paragraph{Baselines}
We consider classical models for tabular data, including GBDT, Logistic Regression, MLP, Sparse MLP \cite{morcos2019one}, and recently proposed TabTransformer \cite{huang2020tabtransformer} and the Variational Information Bottleneck model  \cite{alemi2016deep} as baselines. To ensure a fair and direct comparison, the performance results for all baselines are cited from~\cite{huang2020tabtransformer}, reflecting the outcomes of their original hyperparameter optimization. We adopt the same methodology for ALF, i.e., using 5-fold cross validation splits to train and reporting mean and standard deviation of AUCROC over the 5 splits.

\paragraph{Results}
As shown in Table~\ref{tab:results-public}, ALF outperforms baseline models on three of the five datasets and is competitive on the other two. This advantage was most evident on the large-scale Albert dataset (425,000+ samples, 78 features), where ALF's performance was significantly superior to all other models.

\begin{table}[t!]
\centering
{
\begin{tabular}{c|c|c|c}
\hline 
Dataset Name & Datapoints & Features & Positive Class\%\tabularnewline
\hline 
\hline 
albert & 425240 & 79 & 50\tabularnewline
\hline 
dota2games & 92650 & 117 & 52.7\tabularnewline
\hline 
adult & 34190 & 25 & 85.4\tabularnewline
\hline 
blastchar & 7043 & 20 & 26.5\tabularnewline
\hline 
1995 income & 32561 & 14 & 24.1\tabularnewline
\hline 
\end{tabular}
}
\caption{Benchmark datasets. All datasets are binary classification tasks. Positive Class\% is the fraction of data points that
belongs to the positive class.}
\label{tab:public-data-info}
\end{table}

\begin{table}[t!]
\centering
\resizebox{9cm}{!}
{
\begin{tabular}{c|c|c|c|c|c}
\hline 
Dataset Name & albert & dota2games & adult & blastchar & 1995\_income\tabularnewline
\hline 
\hline 
MLP & 0.740 \textpm{} 0.001 & 0.631 \textpm{} 0.002 & 0.725 \textpm{} 0.010 & 0.839 \textpm{} 0.010 & 0.905 \textpm{} 0.003\tabularnewline
\hline 
Sparse MLP & 0.741 \textpm{} 0.001 & 0.633 \textpm{} 0.004 & 0.740 \textpm{} 0.007 & 0.842 \textpm{} 0.015 & 0.904 \textpm{} 0.004\tabularnewline
\hline 
TabTransformer & 0.757 \textpm{} 0.002 & 0.633 \textpm{} 0.002 & 0.737 \textpm{} 0.009 & 0.835 \textpm{} 0.014 & 0.906 \textpm{} 0.003\tabularnewline
\hline 
TabNet & 0.705 \textpm{} 0.005 & 0.529 \textpm{} 0.025 & 0.663 \textpm{} 0.016 & 0.816 \textpm{} 0.014 & 0.875 \textpm{} 0.006\tabularnewline
\hline 
VIB & 0.737 \textpm{} 0.001 & 0.628 \textpm{} 0.003 & 0.733 \textpm{} 0.009 & 0.842 \textpm{} 0.012 & 0.904 \textpm{} 0.003\tabularnewline
\hline 
Logistic Regression & 0.726 \textpm{} 0.001 & \textbf{0.634 \textpm{} 0.003} & 0.721 \textpm{} 0.010 & 0.844 \textpm{} 0.010 & 0.899 \textpm{} 0.002\tabularnewline
\hline 
GBDT & 0.763 \textpm{} 0.001 & 0.621 \textpm{} 0.004 & \textbf{0.756 \textpm{} 0.011} & 0.847 \textpm{} 0.016 & 0.906 \textpm{} 0.002\tabularnewline
\hline 
ALF & \textbf{0.773 \textpm{} 0.007} & 0.621 \textpm{} 0.005 & 0.733 \textpm{} 0.005 & \textbf{0.848 \textpm{} 0.012} & \textbf{0.911 \textpm{} 0.002}\tabularnewline
\hline 
\end{tabular}
}
\caption{AUC score for models on the benchmark datasets. Values are the mean over 5 cross-validation splits, plus or minus the standard deviation. Larger values mean better result.}
\label{tab:results-public}
\end{table}

\subsection{Evaluation with Production Data}
We deployed ALF to the Google Ads Safety system to identify fraudulent
advertisers who violate various ads policies. Here, we report online
model metrics as well as offline analysis to demonstrate the effectiveness of
our proposed method. 
\subsubsection{Methodology}
\paragraph{Data}
The Google Ads datasets consist of daily
snapshots of advertisers, capturing a comprehensive view of their attributes
and activities at a specific point in time. Each snapshot contains structured
data, text content from ads and landing pages, images, and available video
content. We use daily snapshots to mitigate the noise from the continuous evolution
of advertiser data. The data is split chronologically into training, validation,
and test sets:
\begin{itemize}
    \item Training Set: Over 100 million advertiser snapshots.
    \item Validation Set: Over 10 million snapshots, used for hyperparameter tuning.
    \item Test Set: Over 10 million snapshots from a future timeframe,
      ensuring a realistic evaluation of the model's predictive performance on unseen data.
\end{itemize}

\paragraph{Baselines}
The baseline model considered is our previous production model, which is the
result of an extensive neural architecture search and hyperparameter tuning. The architectures considered 
include DNNs, ensembles, GBDTs, and logistic regression with feature cross
exploration. Notably, the standard architectures within this search space do not support pre-training, focal loss, or calibration via SNGP heads. For each model, the best-performing architecture and hyperparameters are selected automatically via Vizier~\cite{song2024vizier} to ensure optimal performance.

We omit encoder-based models, such as
FLAVA \cite{singh2022flava}, TabTransformer \cite{huang2020tabtransformer}, FT-Transformer\cite{gorishniy2021revisiting}, and SAINT \cite{somepalli2021saint}, from baselines, since they are not designed to handle
the combination of complex structured data, multivalent features, and the
scale required for our application. As shown in Table~\ref{tab:baseline_comparisons},
adapting our dataset to these models would require feature restrictions that would invalidate a fair comparison. We also exclude Large multi-modal generative models (e.g., Flamingo \cite{alayrac2022flamingovisuallanguagemodel}, VLMo \cite{bao2022vlmounifiedvisionlanguagepretraining}) from baselines. They are cost-prohibitive at our scale because they operate on raw data
(e.g., image pixels) instead of embeddings. 

\paragraph{Hyperparameters} We used Vizier \cite{song2024vizier} for hyperparameter optimization. A
selection of the most interesting hyperparameters is presented in Table~\ref{tab:hyperparameters} in Appendix. With
these settings, the total number of trainable parameters for our model is about
322M.

For individual ablation experiments, due to computational constraints, hyperparameters were not re-tuned. However, we observed that the model's
performance was not highly sensitive to minor variations in hyperparameters.

\paragraph{Reporting} Due to the significant computational cost of training, we were unable to perform multiple runs for formal significance testing. However, all reported metrics were computed on a test set of over 10 million examples, which provides a high degree of statistical power for our comparisons.

\subsubsection{Production Metrics}
We successfully deployed our model to a production abuse detection system, where it has led to a significant reduction in policy violations across a range of abuse categories. The system's goal is a binary classification task: identifying advertisers engaged in user harm or financial abuse.

For three key financial abuse policies designated as Policy A, Policy B, and Policy C, as shown in Table~\ref{tab:policy_performance}, our model operates at similar or higher recall values while achieving significantly
higher precision
compared to the baseline production model. 

Our latency as measured in production when running on CPUs is 29ms, 
which is higher than the 8ms observed for the baseline production model. Our model is also larger in parameter count and occupies more CPU memory than 
the baseline production model. However, both of these are within the acceptable 
bounds for the production workflow this model was deployed in. 
Running ALF on TPUs/GPUs should lower the latency for latency-sensitive applications.

\begin{table}[t]
\small
    \centering    
    \begin{tabular}{lcccccc}
        \toprule
        & \multicolumn{2}{c}{\textbf{Policy A}} & \multicolumn{2}{c}{\textbf{Policy B}} & \multicolumn{2}{c}{\textbf{Policy C}} \\
        \cmidrule(lr){2-3} \cmidrule(lr){4-5} \cmidrule(lr){6-7}
        \textbf{Metric} & \textbf{Baseline} & \textbf{ALF} & \textbf{Baseline} & \textbf{ALF} & \textbf{Baseline} & \textbf{ALF} \\
        \midrule
        Precision & 95.3\% & \textbf{99.8\%} & 80\% & \textbf{88\%} & 99.2\% & \textbf{99.5\%} \\
        Recall & 85.7\% & \textbf{92.4\%} & 35\% & \textbf{37\%} & 21.0\% & \textbf{64.0\%} \\
        \bottomrule
    \end{tabular}
    \caption{Performance metrics of ALF vs Baseline across 3 policies in production environment.}
    \label{tab:policy_performance}
\end{table}

\subsubsection{Offline Analysis}

We perform a deep-dive offline analysis on Policy C to benchmark our proposed model against several common tabular architectures. These architectures, shown in Table 1, are the candidates in our extensive neural architecture search and hyperparameter tuning process that resulted in the production baseline model. The evaluation was conducted with two distinct feature sets: one with multi-modal embedding features and one without them. As detailed in Table~\ref{tab:baseline_comparisons}, our proposed model achieves superior AUPRC and AUROC performance compared to all other approaches. The results further reveal that a significant portion of this performance gain is directly attributable to the embedding features, highlighting their critical role in the model's success.

\begin{table}[t!]
\centering
\resizebox{9cm}{!}
{
\begin{tabular}{c|c|c}
\hline 
 & Auprc & Auroc\tabularnewline
\hline 
\hline 
ALF (proposed) & 0.46 & 0.93\tabularnewline
\hline 
ALF without embedding (proposed) & 0.42 & 0.91\tabularnewline
\hline 
Production & 0.41 & 0.91\tabularnewline
\hline 
Production without embedding & 0.45 & 0.74\tabularnewline
\hline 
Logistic Regression & 0.43 & 0.91\tabularnewline
\hline 
Random Forest & 0.40 & 0.90\tabularnewline
\hline 
GBDT & 0.41 & 0.91\tabularnewline
\hline 
Deep Neural Network & 0.38 & 0.87\tabularnewline
\hline 
Regression without embedding & 0.34 & 0.87\tabularnewline
\hline 
Random Forest without embedding & 0.42 & 0.91\tabularnewline
\hline 
GBDT without embedding & 0.45 & 0.91\tabularnewline
\hline 
Deep Neural Network without embedding & 0.35 & 0.88\tabularnewline
\hline 
\end{tabular}

}
\caption{Performance comparison of our proposed model against commonly used methods on Policy C data.}
\label{tab:baseline_comparisons}
\end{table}

In addition, we evaluate our model on the Advertiser Understanding task, which
involves using the embeddings generated by our model to measure the
trustworthiness of Advertisers. As shown in Figure \ref{fig:embeddings},
the outputs of our model are meaningfully separated, supporting our claim that
the embeddings from our model improve performance in tasks focused on
calculating advertiser trustworthiness and clustering fraudulent advertisers.

\subsubsection{Ablation Studies}

\begin{table}[t!]
\centering
\resizebox{9cm}{!}
{
\begin{tabular}{c|c|c|c}
\hline
 & Financial Fraud & User Harm & Average\tabularnewline
\hline
\hline
Structured Features Only & 0.8979 & 0.6262 & 0.7620\tabularnewline
\hline
Content Features Only & 0.4505 & 0.3080 & 0.3792\tabularnewline
\hline
No Spectral Norm & 0.8982 & 0.6260 & 0.7621\tabularnewline
\hline
No Inter-Sample Attention & 0.9075 & 0.6570 & 0.7822\tabularnewline
\hline
No Contrastive Loss & 0.9088 & 0.6568 & 0.7828\tabularnewline
\hline
No Reconstruction Loss & 0.9080 & 0.6581 & 0.7830\tabularnewline
\hline
Our Proposed Model & 0.9102 & 0.6672 & 0.7887\tabularnewline
\hline
\end{tabular}
}
\caption{Ablation experiments on the effects of different components of our
proposed model, evaluated using the AUPR metric.}
\label{tab:ablation}
\end{table}

In Table \ref{tab:ablation}, we present ablation experiments to evaluate the contribution of each component in our proposed method, comparing their effects on a subset of our dataset. The table presents AUPRC performance for financial fraud (e.g., stolen credit cards) and user harm (e.g., phishing websites), with an average in the third column. We observe that the contribution of content features is task-dependent: while structured data provides a strong baseline for financial fraud, content features are critical for user harm as expected, providing a relative AUPRC lift of nearly 7\% (0.6262 to 0.6672). Furthermore, spectrally normalized projections, calibrated probabilistic outputs, Inter-Sample Attention, Contrastive Loss, and Reconstruction Loss are identified as crucial components, each significantly contributing to the model's overall performance.

We intentionally omit ablations on the choice of 'K' and
the specific criteria for top-K asset selection. Revealing these details could
provide adversaries with insights into our enforcement strategies, potentially
enabling them to circumvent our detection systems.

\subsubsection{Embedding Quality Analysis}

\begin{figure}[t]
    \centering
    \includegraphics[width=0.6\linewidth,scale=1]{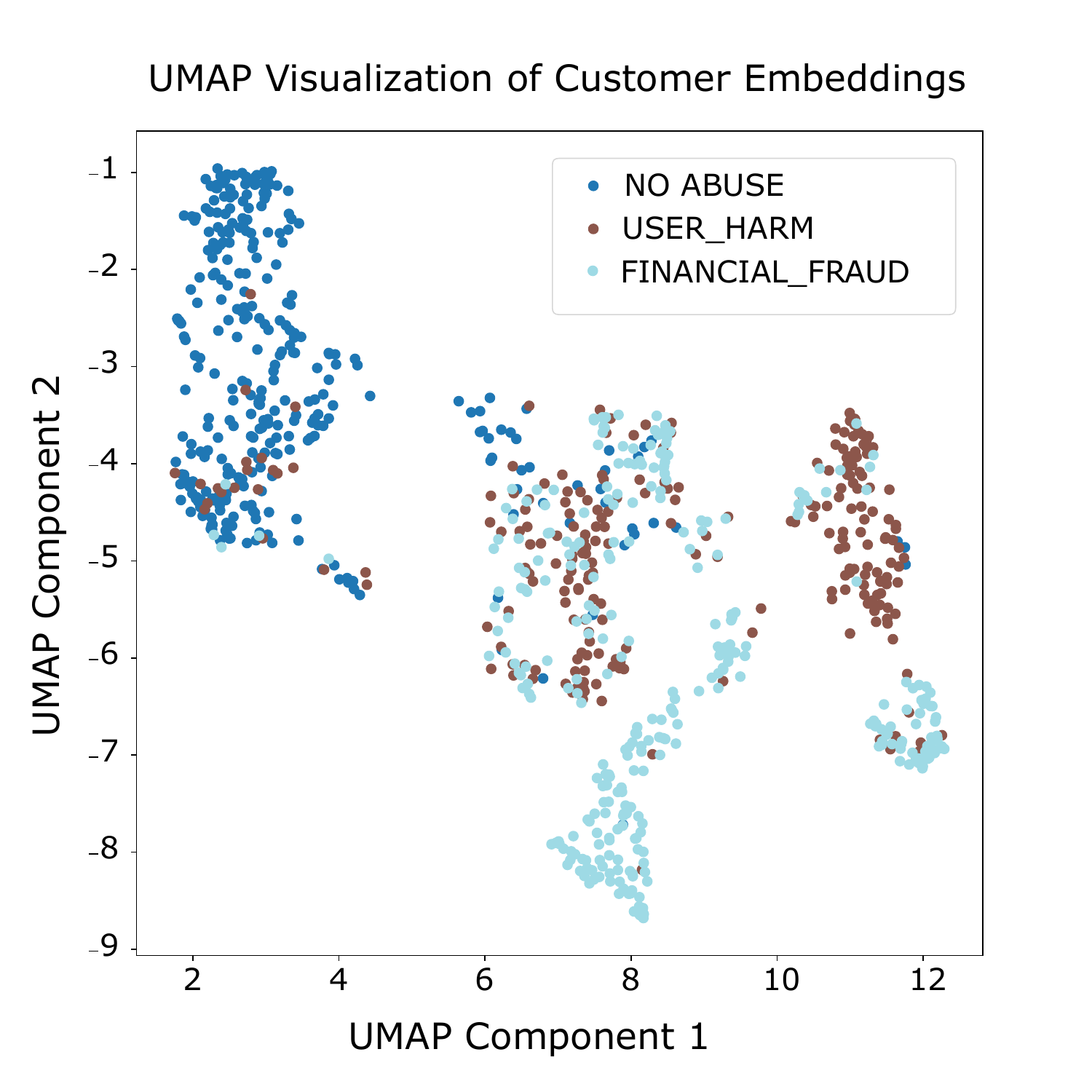}
    \caption{UMAP visualization of advertiser embeddings, colored by advertiser
    intent.}
    \label{fig:embeddings}
\end{figure}
\begin{figure}[t]
    \centering
    \includegraphics[width=0.6\linewidth,scale=1]{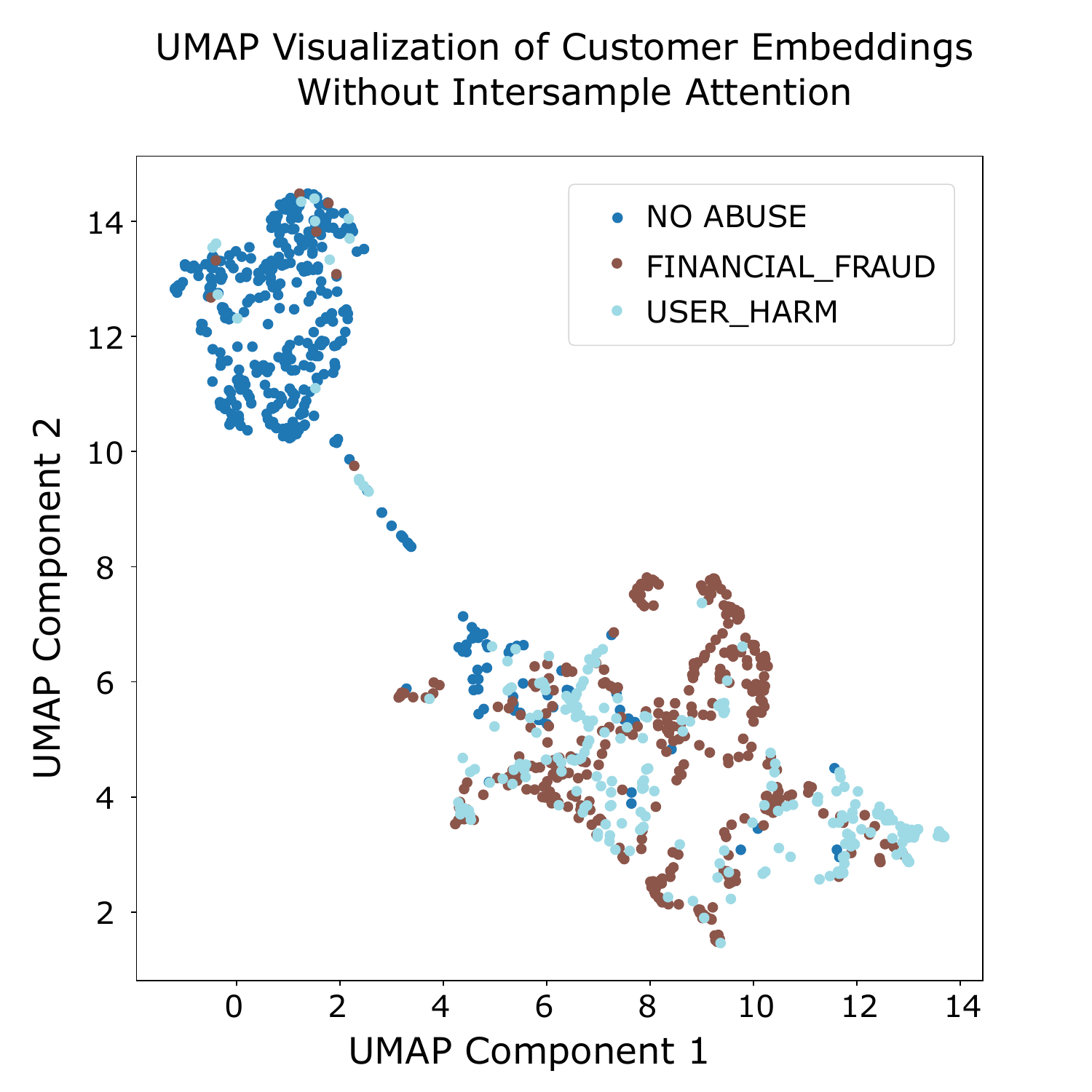}
    \caption{UMAP visualization of advertiser embeddings \emph{without}
    intersample attention, colored by advertiser intent.}
    \label{fig:simple_embeddings}
\end{figure}

Figure \ref{fig:embeddings} shows the learned embeddings clearly clustered by
advertiser intent in our final model. In contrast, Figure
\ref{fig:simple_embeddings} shows the embeddings from a lower-quality model
when we remove the inter-sample attention, where we can see that the
Advertiser Intents are not well-clustered or separated. We observe a noticeable
difference especially when it comes to separating user harm policies from
financial fraud.

\section{Conclusion}
\label{sec:conclusion}
This paper addresses a critical gap between two lines of research: large multi-modal models, which are often impractical for predictive tasks on structured data due to their prohibitive cost at scale, and tabular models that cannot natively handle multi-modal signals. We bridge this gap through ALF, a framework that demonstrates the successful real-world deployment of a state-of-the-art transformer architecture for large-scale advertiser understanding. ALF is designed to directly solve the key industrial challenges of heterogeneous data, unbounded creative assets, and the need for real-world reliability. To tackle heterogeneous data, it uses an early-fusion, dual-attention mechanism; the inter-sample attention component is made computationally feasible at our scale through a scalable, projection-based approach we introduced, overcoming key limitations of prior work. For unbounded assets, it employs an efficient top-k selection on embeddings. To ensure reliability, it integrates SNGP heads for well-calibrated predictions.

The effectiveness of this approach was validated against an exceptionally strong production baseline, itself the result of an extensive search across various architectures and hyperparameters, including DNNs, ensembles, GBDTs, and logistic regression with feature cross exploration. While ALF’s latency is higher due to its larger model size, it remains well within the acceptable range for our production environment and can be further optimized using hardware accelerators. Experiments show ALF significantly outperforms the baseline on key risk detection tasks, a performance lift driven by its unique ability to holistically model content embeddings, which simpler architectures struggled to leverage. This trade-off is justified by its successful deployment, where ALF serves millions of requests daily.

Future work could explore several promising directions. The current model could be extended to incorporate temporal dynamics to better detect evolving abuse patterns. Additionally, investigating scaling properties and simplifying the architecture to reduce complexity offers a valuable avenue for analyzing model size-performance trade-offs and broadening applicability. While our top-k approach for handling unbounded assets works well in practice, a theoretical analysis of its information loss could guide further improvements. Finally, the success of our unified architecture suggests its potential for a broader range of advertiser-related tasks, including creative optimization and audience modeling. In conclusion, ALF not only provides an effective solution for multi-modal entity modeling but also serves as a valuable case study for deploying advanced transformer architectures in complex, high-stakes industrial systems.

\section{Acknowledgements}
\label{sec:acknowledgements}

We extend our gratitude to Anish Das Sarma for funding and providing initial
feedback for this effort. We thank Benoît Raybaud and Nihar Khedekar for
leading the team that made this work possible. Special thanks to Dongjin Kwon,
Alborz Mazloomian, Jossue Loubet, and Mike Martin for their tireless efforts
in feature processing and for successfully bringing the model into production. 
We also thank Shengwei Xu for his insightful discussions and review of the manuscript.
Finally, we acknowledge the invaluable support from Jun Gao’s and Dake He’s
teams in bridging the gap between modeling and real-world business impact.

\bibliographystyle{ACM-Reference-Format}
\bibliography{references}

\appendix
\balance
\section{Responsible AI Considerations}
The development and deployment of ALF were guided by responsible AI principles
to mitigate risks and ensure fair, transparent operation.

\textbf{Mitigating Misuse:} The model is used exclusively within Google's
internal systems for advertiser risk assessment and is not publicly released.
This controlled environment minimizes the potential for external misuse.

\textbf{Privacy:} All advertiser data is processed after being stripped of
Personally Identifying Information (PII), ensuring that the model does not rely
on sensitive user data.

\textbf{Human Oversight and Accountability:} The model's predictions are not
used in isolation. They serve as a signal within a larger system that includes
robust human review and appeals processes. False positives are carefully
monitored, and advertisers have channels to appeal decisions, ensuring
accountability and fairness.

\textbf{Combating Adversarial Behavior:} A key challenge in this domain is the
presence of adversaries who actively try to conceal fraudulent or malicious
assets. ALF's architecture is designed to enhance the effectiveness of our
existing asset aggregation and selection techniques. By processing a holistic
representation of an advertiser, the model can better identify subtle
inconsistencies and suspicious patterns that might be missed when analyzing
assets in isolation. While specific details of our anti-abuse strategies are
omitted to prevent them from being circumvented, the model's design plays a
crucial role in our resilience to adversarial attacks.
\section{Additional Details for Experiments}
We present links to datasets and hyperparameters of our model in the tables below.

\begin{table}[H]
\centering
\resizebox{8.8cm}{!}
{
\begin{tabular}{lcc}
\toprule
\textbf{Hyperparameter} & \textbf{Value} \\
\midrule
Batch size & 32768 \\
Dimension of the created embedding & 32 \\
Number of attention heads in the transformer & 8 \\
Number of layers in the transformer & 6 \\
Dimensionality of the intermediate layer in the transformer & 512 \\
Activation & gelu \\
Optimizer & AdamW \\
Learning rate schedule & cosine decay \\
Initial learning rate (pretrain and finetune) & 5e-05 \\
Learning rate cosine decay alpha (pretrain and finetune) & 0.1 \\
Learning rate decay steps factor (pretrain and finetune) & 0.95 \\
Learning rate warm up target (pretrain) & 10 \\
Learning rate warm up target (finetune) & 2 \\
\bottomrule
\end{tabular}}
\caption{Best hyperparameter values for production data}
\label{tab:hyperparameters}
\end{table}

\begin{table}[H]
\centering
\resizebox{9cm}{!}
{
\begin{tabular}{c|c|c}
\hline 
Dataset Name & URL & \tabularnewline
\hline 
\hline 
albert & http://automl.chalearn.org/data & \tabularnewline
\hline 
dota2games & https://archive.ics.uci.edu/ml/datasets/Dota2+Games+Results & \tabularnewline
\hline 
adult & http://automl.chalearn.org/data & \tabularnewline
\hline 
blastchar & https://www.kaggle.com/blastchar/telco-customer-churn & \tabularnewline
\hline 
1995 income & https://www.kaggle.com/lodetomasi1995/income-classification & \tabularnewline
\hline 
\end{tabular}

}
\caption{Benchmark Dataset Links.}
\label{tab:public-data-links}
\end{table}

\end{document}